\documentclass[letterpaper, 10 pt, conference]{ieeeconf}  

\IEEEoverridecommandlockouts                              

\usepackage{graphicx} 
\usepackage{epstopdf}
\DeclareGraphicsExtensions{.eps}

\usepackage{multirow}
\usepackage{algorithmic}
\usepackage{algorithm}
\usepackage{amsmath} 
\usepackage{amssymb}
\usepackage{amsxtra}
\usepackage{url} 
\usepackage{color} 
\usepackage{bm}
\usepackage{placeins}
\usepackage[caption=false]{subfig}
\usepackage{xspace}
\usepackage{rotating}
\usepackage{siunitx}
\usepackage{textcomp}%

\newcommand{\eg}{e.g.,\xspace}

\newcommand{\wrt}{\text{w.r.t.}\xspace}

\newcommand{\myvector}[1]{\bm{#1}}
\newcommand{\myvec}[1]{\myvector{#1}}

\newcommand{\R}[1]{\mathbb{R}^{#1}}

\newcommand{\algorithmicinput}{\textbf{Input:}}
\newcommand{\algorithmicoutput}{\textbf{Output:}}
\newcommand{\algorithmicproperty}{\textbf{Property:}}
\newcommand{\INPUT}{\item[\algorithmicinput]}
\newcommand{\OUTPUT}{\item[\algorithmicoutput]} 
\newcommand{\PROPERTY}{\item[\algorithmicproperty]}

\newcommand{\excise}[1]{}

\newif\ifremark
\long\def\remark#1{
  \ifremark%
  \begingroup%
  \dimen0=\textwidth
  \advance\dimen0 by -1in%
  \setbox0=\hbox{\parbox[b]{\dimen0}{\protect\em #1}}
  \dimen1=\ht0\advance\dimen1 by 2pt%
  \dimen2=\dp0\advance\dimen2 by 2pt%
  \vskip 0.25pt%
  \hbox to \textwidth{%
    \vrule height\dimen1 width 3pt depth\dimen2%
    \hss\copy0\hss%
    \vrule height\dimen1 width 3pt depth\dimen2%
  }%
  \endgroup%
  \fi}

\newcommand{\x}{\ensuremath{\myvec{s}}}

\newcommand{\uv}{\ensuremath{\myvec{a}}}
\newcommand{\ui}[1]{a}

\newcommand{\ac}{\uv}

\title{\LARGE \bf
Learning Navigation Behaviors End-to-End with AutoRL\\
}

\author{Hao-Tien Lewis Chiang$^{1,*}$, Aleksandra Faust$^{1,*}$, Marek Fiser$^{1}$, Anthony Francis$^{1}$
\thanks{$^{1}$Google AI, Mountain View, CA 94043, USA {\tt\scriptsize {lewispro,faust,mfiser,centaur}@google.com}}%
\thanks{$^{*}$ Authors contributed equally.}%
}

\begin{document}

\maketitle
\thispagestyle{empty}
\pagestyle{empty}

\begin{abstract}
We learn end-to-end point-to-point and path-following navigation behaviors that avoid moving obstacles. 
These policies receive noisy lidar observations and output robot linear and angular velocities. The policies are trained in small, static environments with AutoRL, an evolutionary automation layer around Reinforcement Learning (RL) that 
searches for a deep RL reward and neural network architecture with large-scale hyper-parameter optimization. AutoRL first finds a reward that maximizes task completion, and then finds a neural network architecture that maximizes the cumulative of the found reward. Empirical evaluations, both in simulation and on-robot, show that AutoRL policies do not suffer from the catastrophic forgetfulness that plagues many other deep reinforcement learning algorithms, generalize to new environments and moving obstacles, are robust to sensor, actuator, and localization noise, and can serve as robust building blocks for larger navigation tasks. Our path-following and point-to-point policies are respectively 23\% and 26\% more successful than comparison methods across new environments. Video at: \url{https://youtu.be/0UwkjpUEcbI}.
\end{abstract}

\section{INTRODUCTION}

\begin{figure}[tb]
    \centering
    \begin{tabular}{cc}
    \subfloat[Narrow corridor]{\includegraphics[trim=0mm 0mm 15mm 0mm,clip,width=0.22\textwidth, height=4cm,keepaspectratio=false]{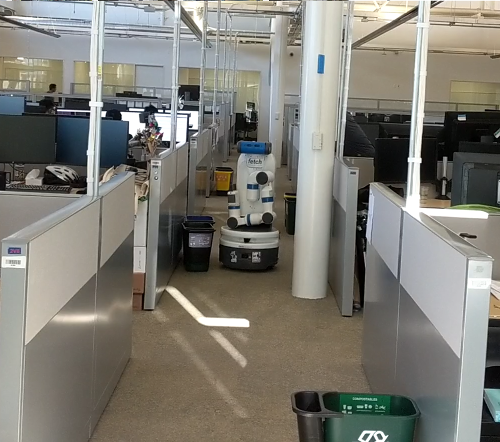}\label{fig:fetchNarrow}}&
    \subfloat[Unstructured dynamic]{\includegraphics[trim=40mm 0mm 70mm 0mm,clip,width=0.22\textwidth,height=4cm,keepaspectratio=false]{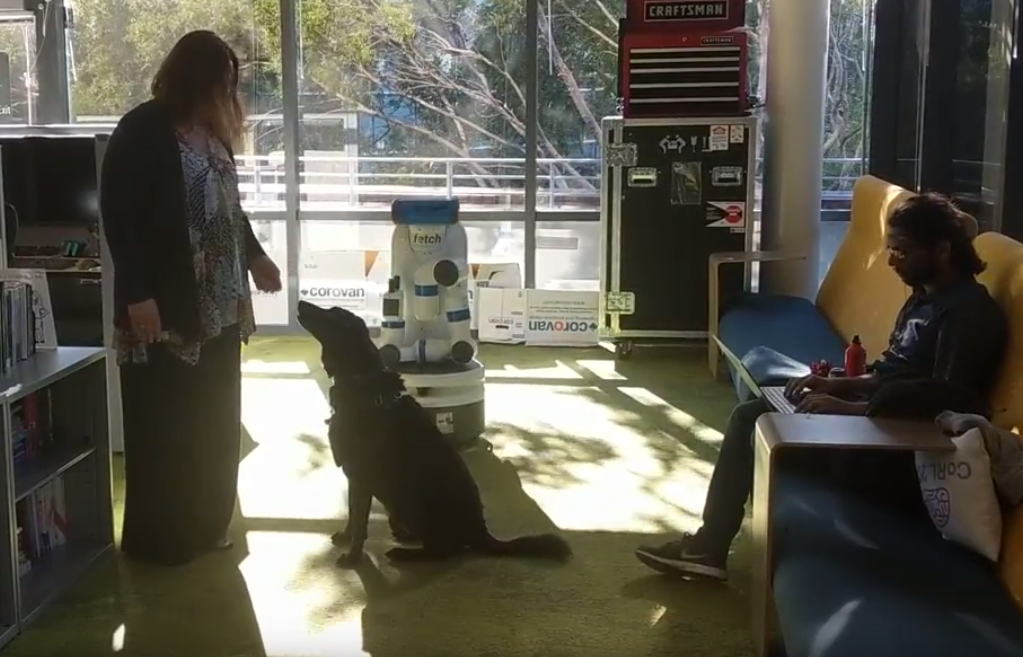}\label{fig:fetchUnstructured}}
    \end{tabular}
    \caption{\small Navigation behaviors on a Fetch: (a) path following in a narrow corridor (b) point to point in a dynamic environment. \label{fig:fetchFun}}
\end{figure}

Assistive robots, last-mile delivery, warehouse navigation, and robots in offices require robust robot navigation in dynamic environments (Fig. \ref{fig:fetchFun}). 
While good methods exist for robot navigation sub-tasks such as 
localization and mapping, motion planning, and control, current local navigation methods typically must be tuned for each new robot and environment \cite{chen2015deepdriving}. For example, vision-based navigation is robust to noisy sensors but typically relies on high-level motion primitives, such as ``go straight'' and ``turn left'' that abstract away robot dynamics \cite{mousavi_eccv18}.
Robot navigation requires behaviors that are useful to existing navigation stacks, easily transfer from simulation to physical robots or new large-scale environments with minimum tuning, and robustly avoid obstacles despite noisy sensors and actuators.

We define robot navigation behaviors as intelligent agents that depend on the robot's sensors, actuators, dynamics and geometry, without relying on foreknowledge of its environment \cite{antonelo2015learning}. Many navigation tasks can be accomplished using two basic behaviors: local planning, i.e., point to point, which creates trajectories from the robot's current configuration to a given target \cite{prm-rl}, and path following, which stays near a given guidance path \cite{chiang2015path}. Both behaviors need to produce dynamically feasible trajectories that are robust to many kinds of noise and inaccuracies, rely only on observations coming from primitive sensors, and avoid unexpected obstacles. In this paper, we present a reliable method to implement these navigation behaviors by learning end to end polices that directly map sensors to controls, and we show that these policies transfer from simulation to physical robots and new environments while robustly avoiding obstacles. 

\begin{figure*}[ht]

\begin{tabular}{cccc}
    \subfloat[Training, 23 by 18m]{\includegraphics[trim=0mm 0mm 0mm 0mm,clip,width=0.17\textwidth,keepaspectratio=true]{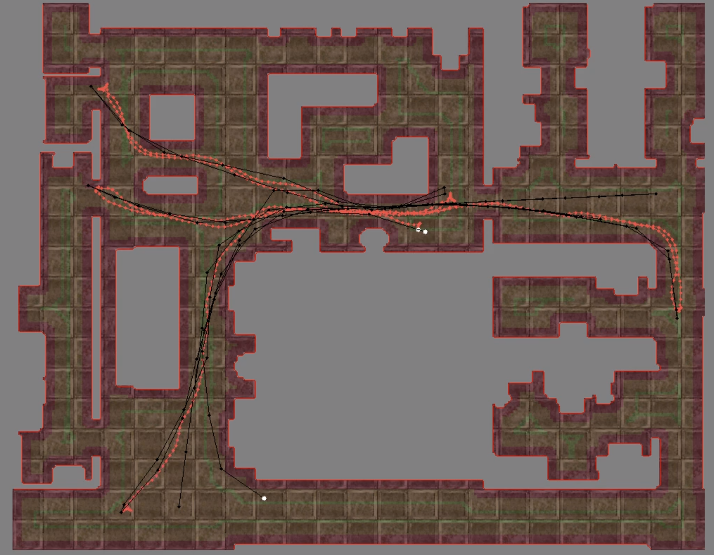}\label{fig:trainingEnv}}&
    \subfloat[Building 1, 183 by 66m]{\includegraphics[trim=0mm 0mm 0mm 0mm,clip,width=0.37\textwidth,keepaspectratio=true]{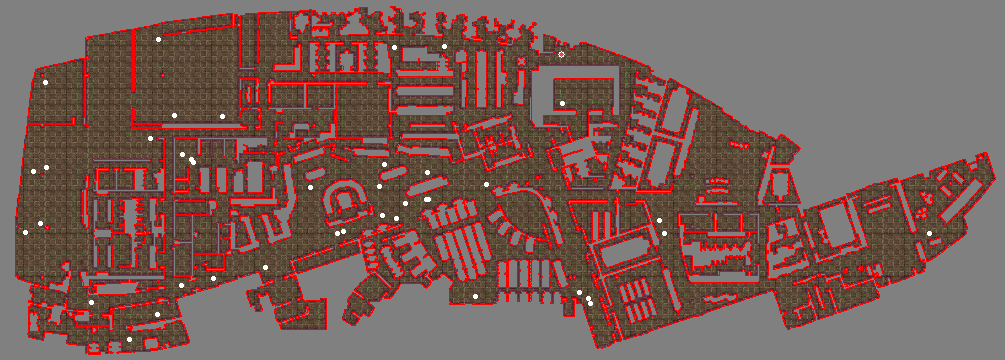}\label{fig:mtv40}}&
    \subfloat[Building 2, 60 by 47m]{\includegraphics[trim=0mm 0mm 0mm 0mm,clip,width=0.17\textwidth,keepaspectratio=true]{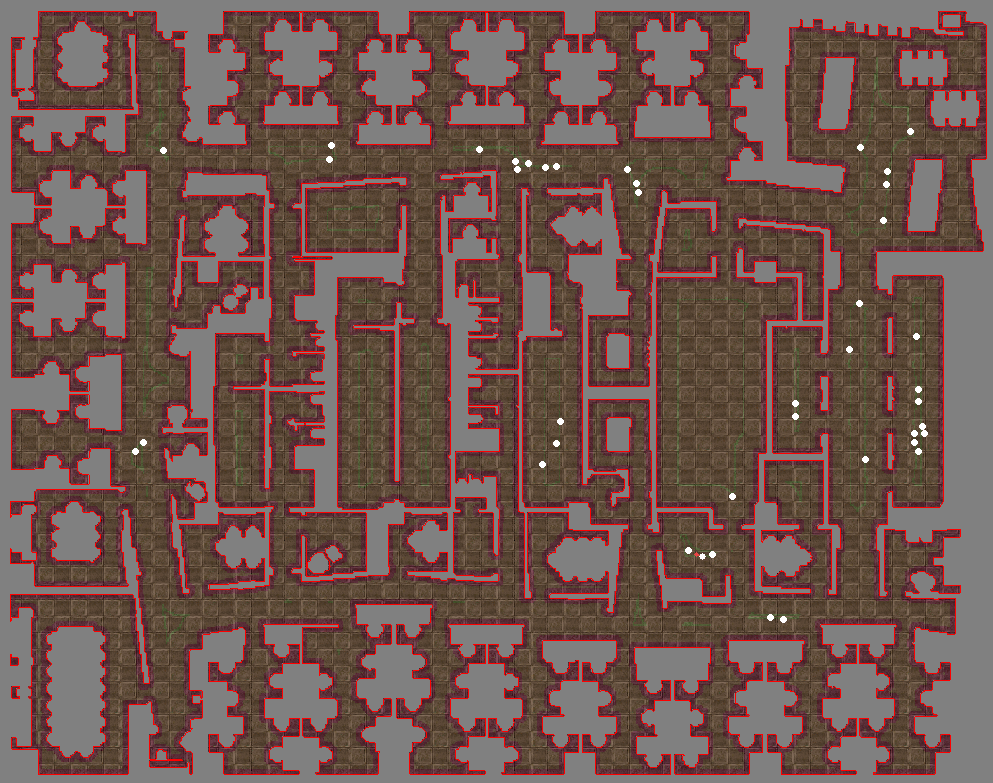}\label{fig:sfo1ms}}&
    \subfloat[Building 3, 134 by 93m]{\includegraphics[trim=0mm 0mm 0mm 0mm,clip,width=0.19\textwidth,keepaspectratio=true]{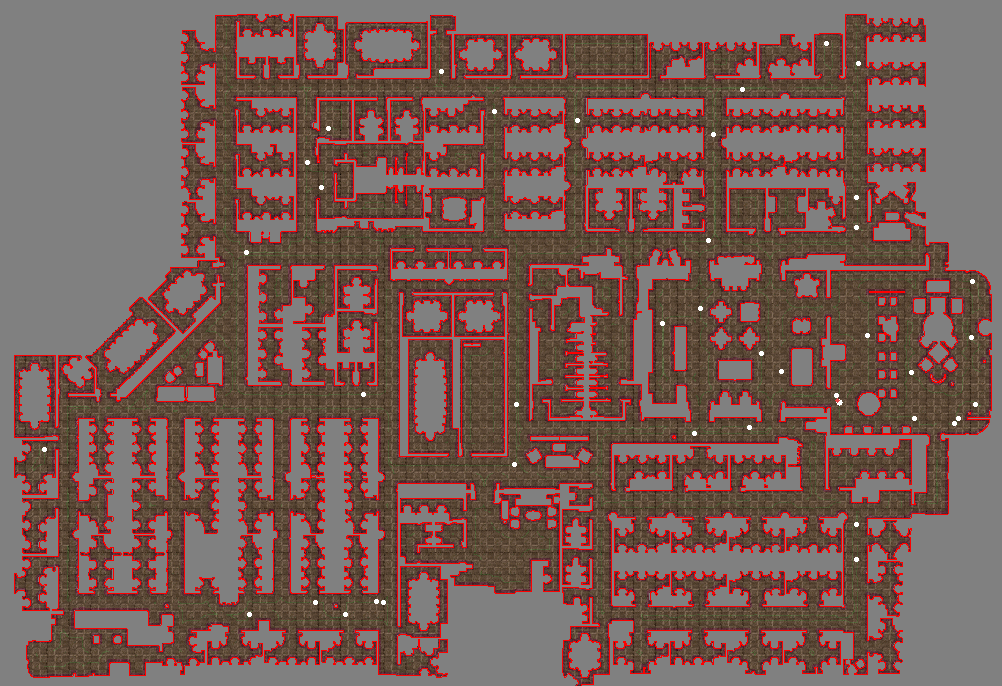}\label{fig:mtv1667}}
\end{tabular}
    \caption{\small (a) Training and (b, c, d) evaluation environments with 40 moving obstacles (white circles). The evaluation environments are 6.8 to 30.1 times bigger than the training, generated from real office building plans. \label{fig:environments}}
\end{figure*}

To do so, we rely on deep Reinforcement Learning (RL), which has shown great progress in visual navigation \cite{vin} and video games \cite{atari-paper}. RL learns policies that map between observations and controls to maximize the cumulative return. Deep RL has enabled learning tasks that have been difficult or impossible to program manually, from matching human performance \cite{atari-paper} to locomotion of humanoid figures \cite{dm-curriculum}. 
Despite the promise of deep RL, training these policies is difficult and requires careful consideration when selecting the reward function and choosing the neural network architecture. Specifically, training even simple tasks can fail if rewards are sparse - that is, if the success conditions of the true objective are hard to discover \cite{montezumaRevenge}; this is true for navigation in large spaces. Reward shaping \cite{reward_shaping,reward_shaping_robots} addresses this problem by introducing a proxy reward function that is less sparse than the true objective. However, poorly chosen shaped rewards can lead to pathologies where agents learn to exploit the reward function or to terminate the episode early \cite{ai-death}. Designing good reward functions is well understood for some areas (\eg video games \cite{atari-paper}), but for most tasks, including navigation, it remains a black art. Similarly, selection of neural network architecture often consists of trial and error, and remains another source of training difficulties.

Hyperparameter tuning can improve learning behavior: 
gradient-free, evolutionary methods, such as Covariance Matrix Adaptation Evolution Strategy (CMA-ES) \cite{cmaes}, can transform an algorithm that fails to converge into one that produces near-optimal policies. 
To simplify deep RL training, we present AutoRL, which automates the search for both the shaped reward and the neural network architecture. AutoRL combines deep RL with gradient-free hyperparameter optimization. It takes a parameterized reward function along with the true objective, and then uses large-scale hyperparameter optimization to shape the reward. With this reward fixed, it then optimizes network layer sizes to identify the most successful policies for the given task. Applied to point-to-point and path-following tasks, AutoRL learns robust policies that cope with both static and dynamic obstacles, even though the tasks are trained in simulation in small environments with only static obstacles. 
The specific contributions of this work are the AutoRL method as well as point-to-point and path-following navigation policies.

We evaluate these policies in three large building environments (Fig. \ref{fig:environments}) with respect to various noise levels and a number of moving obstacles. Our simulation results show the path following and point to point policies have higher success rate in novel environments, on average 23\% and 26\% higher over the baseline, respectively, and are robust to noise. In experiments on a real robot (Fig. \ref{fig:fetchFun}), we observe reliable obstacle avoidance in dynamic environments and collision-free navigation along 80+ m paths. 

\section{RELATED WORK}
\label{sec:related-work}

\textit{Motion planning:} Practical robot navigation typically uses a layered navigation stack: given an occupancy map generated by a mapping algorithm such as SLAM \cite{slam}, a global planner such as PRM \cite{kavraki-prm} or RRT \cite{rrts} finds a near-optimal path that
a local planner (\eg \cite{dwa}) executes. 
This ``stacked'' approach has demonstrated reliable operation over long distances \cite{office-marathon}.
Typical planners for dynamic environments include Artificial Potential Field (APF), velocity obstacle, and sampling-based methods \cite{chiang2017safety}. Many of those methods, such as DRT \cite{chiang2015path} or PEARL \cite{faust2016avoiding}, require perfect knowledge of obstacle position or dynamics, which is difficult to obtain when the robot has limited and noisy sensors, like 1D lidar. Like these approaches, we assume reasonably accurate localization, but we couple sensing directly with controls to remove the need for knowing obstacle location or dynamics.

\textit{Deep learning for navigation:} Machine learning for navigation has seen a recent boom, particularly in point-to-point navigation with vision \cite{mousavi_eccv18,vin}.
Some approaches discretize state and action spaces to enable simple RL algorithms \cite{altuntacs2016reinforcement}, whereas others learn low-level controllers end to end using deep learning \cite{chen2015deepdriving}.
Our prior work on PRM-RL uses a point-to-point policy trained with DDPG as the local planner for PRM \cite{prm-rl}. 
Like these approaches, we map from sensors to velocity control, but we train with AutoRL, which tunes the agent's reward and network architectures to improve the policy performance and time needed for hand-engineering. 
Learning by optimizing multiple objective functions, rules, or perceptual features has proven effective for variants of the navigation problem
\cite{fathinezhad2016supervised}; 
other work uses hierarchical RL to break the problem apart into tractable components
\cite{bischoff2013hierarchical}. 
Unlike these methods, we automate the tuning by learning the proxy reward function and neural network architecture instead of imposing it.
Some recent work features learning based sensor to controls navigation in the presence of moving obstacles. For example, Intention-Net \cite{intention-net} maps camera observations and intents to controls, while Pfeiffer et al. \cite{Pfeiffer2017FromPT} navigate from lidar inputs. Both of those methods require demonstrations for training, while AutoRL learns by reinforcement.

\textit{Reinforcement learning with sparse rewards:}
Three main approaches handle sparse rewards in RL: curriculum learning, bootstrapping, and reward shaping. Curriculum methods make tasks progressively more difficult during training, e.g., by moving the goal further from the start in navigation \cite{reverse-currriculum,back-curriculum}. In contrast, AutoRL does not change task difficulty during training. Bootstrapping allows RL policies to be initialized from either hand-engineered polices or recorded episodes \cite{smart2002effective}. AutoRL starts with an uninitialized policy and trains it from scratch. Reward shaping learns from a parameterized dense reward \cite{ng1999policy}. AutoRL is a reward-shaping method. While reward shaping can be done with curriculum and bootstrapping \cite{back-curriculum, smart2002effective}, AutoRL uses hyperparameter optimization rooted in evolutionary algorithms to accomplish this. Further, curriculum learning and bootstrapping can be applied in addition to the hyperparameter optimization.

\textit{Hyper-parameter optimization in deep learning:}
AutoML methods learn the neural network architecture for a given problem. AutoML achieves this with gradient descent \cite{andrychowicz2016learning}, reinforcement learning \cite{wang2016learning}, and even hyper-parameter optimization \cite{real2017large}. AutoRL extends the AutoML approach to RL and subjects both the network architecture and the reward function to  hyper-parameter optimization.


\section{METHODS}
\label{sec:methods}

To learn point-to-point and path-following navigation behaviors, we choose continuous reinforcement learning with partial state observations, which we model with Partially Observable Markov Decision Process (POMDP). We model the agent as an $(O, A, D, R, \gamma)$ tuple representing a partially observable Markov decision process (POMDP) with continuous observations and actions. Observations, actions, and dynamics are determined by the robot and are common to both tasks; we describe them in Section \ref{sec:pomdp}. Point to point and path following share observations, actions, and dynamics, but have different objectives, that are modeled through rewards in POMDP. Sections \ref{sec:method:p2p} and \ref{sec:method:pf} describe the true task objectives, specific rewards, and additional observations specific to each behavior. Section \ref{sec:prl} describes the hyper-parameters of a reinforcement learning algorithm, and Section \ref{sec:ddpg} presents the AutoRL algorithm that learns both the hyperparameters and the policy.

\subsection{POMDP Setup}
\label{sec:pomdp}
The observations, $\myvec{o} = (\myvec{o}_l,\, \myvec{o}_g)^{\theta_n} \in O,$ are ${\theta_n}$ pairs of 1-D lidar vectors, $\myvec{o}_l,$ and goal set, $\myvec{o}_g,$ observed over the last $\theta_n$ steps. 
The agent is a unicycle robot and is controlled by a 2-dimensional continuous action vector, $\ac = (v, \phi) \in A,$ that encodes the robot's linear and angular velocity. The dynamics, $D$, is encoded in the simulator or implicit in the real world.
The remaining factors encode the task: $\gamma \in (0,1) $ is a scalar discount factor, whereas the structure of the reward R is more complicated and we discuss it next. 

The agent's goal is to complete a true navigation objective. For point-to-point navigation this is arriving at a goal location, while for path following this is traversing the entire path by reaching all its waypoints. We can formalize this problem as learning a policy that maximizes the probability of reaching the true objective, $G$,
\begin{equation}
\label{eq:obj}
\myvec{\tilde\pi} = \arg \max_{\myvec{\pi}} \mathbb{P}(G(\x)|\myvec{\pi}), 
\end{equation}
where $\mathbb{P}(G(\x)|\myvec{\pi})$ means that true objective $G$ is reachable from the state $\x$ under control of policy $\myvec{\pi}.$ At the same time, RL learns a policy that maximizes the cumulative reward. We could use the true objective as a reward, but it is sparse, and there are other requirements, such as dynamical feasibility, smooth trajectories, avoiding obstacles, and sensory/motor limitations. We can formulate these requirements as parameterized atomic rewards which provide more timely feedback to aid learning. To that end we represent rewards with
\begin{equation}
\label{eq:reward}
R_{\myvec{\theta_{r}}}(\x, \ac) = \sum_{i=1}^{n_r}\myvec{r}_i(\x, \ac, \myvec{\theta_{r_i}}),    
\end{equation}
where $\myvec{r}_i(\x, \ac, \myvec{\theta_{r_i}})$ is a parameterized atomic reward, and $\myvec{\theta_{r}} = [\myvec{\theta_{r_1}} \cdots \myvec{\theta_{r_{n_r}}}]$ becomes a hyperparameter to be tuned.

\subsection{Point-to-Point (P2P) Task}
\label{sec:method:p2p}
The goal of the P2P behavior is to navigate a robot to a goal position without collision. We assume the robot is well-localized using traditional methods.
The P2P behavior can be used as a local planner for sampling-based planners \cite{prm-rl} in order to navigate large environments. We require the agent to navigate to goals that are not within clear line of sight, but not far enough to require higher-level knowledge of the environment. For example, we expect it to navigate around a corner, but not to a room in a maze of hallways. 

The true objective of P2P is to maximize the probability of reaching the goal during an episode,
\begin{equation}
\label{eq:p2p_obj}
G_{\text{P2P}}(\x) = \mathbb{I} (\|\x - \x_g \| < d_{\text{P2P}}),
\end{equation}
where $\mathbb{I}$ is an indicator function, $\x_g$ is the goal pose, and $d_{\text{P2P}}$ the goal size. The goal observation $\myvec{o}_g$ is the relative goal position in polar coordinates, which is readily available from localization. 
The reward for P2P is:
\begin{equation}
R_{\myvec{\theta_{r_\text{P2P}}}} = 
\myvec{\theta_{r_\text{P2P}}}^T [r_\text{step} \, r_\text{goalDist} \, r_\text{collision} \, r_\text{turning} \, r_\text{clearance} \, r_\text{goal}],
\label{eq:p2p_reward}
\end{equation}
where $r_\text{step}$ is a constant penalty step with value 1, 
$r_\text{goalDist}$ is the negative Euclidean distance to the goal,
$r_\text{collision}$ is 1 when the agent collides with obstacles and 0 otherwise,
$r_\text{turning}$ is the negative angular speed,
$r_\text{clearance}$ is the distance to the closest obstacle, and
$r_\text{goal}$ is 1 when the agent reaches the goal and 0 otherwise. 

\subsection{Path-Following (PF) Task}
\label{sec:method:pf}
The goal of the PF behavior is to follow a guidance path represented by a sequence of waypoints in the workspace. The collision-free guidance path does not need to be dynamically feasible, and can be generated by off-the-shelf path planners like PRMs or A*, or even manually. 
In navigation stack terms, PF is trajectory tracking.

Guidance paths for real-world navigation are often long (100+ m) and have many waypoints with varied separation. The varied input size and non-uniformity pose a challenge for RL methods using neural networks.
To address this problem, we augment the original guidance path $\mathcal{P}_o$ with intermediate waypoints that are linearly interpolated with a constant separation $d_\text{ws}$ between consecutive waypoints. The result is a new guidance path $\mathcal{P}$ consisting of both original and new waypoints. The separation, $d_\text{ws}$, between waypoints is a hyper-parameter that AutoRL searches for.
The waypoint $\myvec{w}_i$ is considered reached iff 1) the previous waypoint $\myvec{w}_{i-1}$ is already reached and 2) the robot is within $d_\text{wr}$ of $\myvec{w}_i$.

The true objective of PF is to reach as many waypoints per episode as possible:
\begin{equation}
\label{eq:pf_obj}
G_{\text{PF}}(\x) = \frac{\sum_{\myvec{w} \in \mathcal{P}}\mathbb{I} (\|\x - \myvec{w} \| < d_\text{wr})}{\|\mathcal{P}\|}.
\end{equation}
where $\mathbb{I}$ is an indicator function that returns 1 if waypoint $\myvec{w}$ is reached and 0 otherwise.
The goal observation of PF, $\myvec{o}_g$, is a partial path consisting of the $N_\text{partial}$ un-reached waypoints.
The reward for PF is:
\begin{equation}
R_{\myvec{\theta_{r_\text{PF}}}} = 
\myvec{\theta_{r_\text{PF}}}^T [r_\text{step} \ r_\text{dist} \ r_\text{collision} \ r_\text{clearance}].
\label{eq:pf_reward}
\end{equation}
where $r_\text{step}$ is a penalty step with value 1. 
$r_\text{dist}$ is the Euclidean distance to the first un-reached waypoint. 
$r_\text{collision}$ is 1 when the agent collide with obstacles and 0 otherwise.
$r_\text{clearance}$ is a penalty with value 1 when the the agent is within $d_{\text{clearace}}$ m away from obstacles and 0 otherwise.

\subsection{Reinforcement Learning Parametrization}
\label{sec:prl}

With observations, action space, true objectives, and parameterized rewards defined, training deep RL requires selecting neural network architecture. Network architecture affects the quality of the trained agent: the capacity of the network determines what the agent can learn. In this work, we use feed-forward fully-connected networks and fix network depth, leaving the size of each layer as our tunable hyperparameter.
However, the hyperparameter tuning technique can be applied to any network architecture.

Let $FF(\theta),$ for $\theta \in \R{n}$ be a feed-forward fully-connected neural network with RELU units and $n$ layers, where the $i^{th}$ layer contains $\theta_i$ neurons. Let us denote the learnable weights of the feed-forward network $FF(\theta)$ with $W_{\theta}.$
Let 
$$ RL(Actor(\myvec{\theta}_{\myvec{\pi}}), Critic(\myvec{\theta_{Q}}), R(\myvec{\theta}_{\myvec{r}}))$$
be a parameterized actor-critic reinforcement learning algorithm that learns policy $\myvec{\pi}(\x | W_{\myvec{\pi}}) = FF(\myvec{\theta}_{\myvec{\pi}})$ and critic $Q(\x, \uv | W_{Q}) = FF(\myvec{\theta}_{Q})$. Actor and critics network architectures are parameterized with $\myvec{\theta}_{\myvec{\pi}}, \myvec{\theta}_Q,$ and reward function, $R$ given in Eq. \eqref{eq:reward}, parameterized with vector $\myvec{\theta}_{\myvec{r}}.$ Let $Obj(\myvec{\theta}_{\myvec{\pi}}, \myvec{\myvec{\theta_{Q}}}, \myvec{\theta}_r | G) \in \R{}$ be the true objective the trained actor $\myvec{\pi}(\x | W_{\myvec{\pi}})$ achieved for the corresponding
$RL(Actor(\myvec{\theta}_{\myvec{\pi}}), Critic(\myvec{\theta_{Q}}), R(\myvec{\theta}_{\myvec{r}}))$.

\subsection{AutoRL}
\label{sec:ddpg}

We automate RL hyperparameter selection with the AutoRL evolutionary search procedure, described in Algorithms \ref{alg:autorl} and \ref{alg:hp}. We split shaping in two phases, reward shaping (Line 1, Alg. 1) and network shaping (Line 2), because the search space grows exponentially in the number of tuning parameters. First, we find the best reward function \wrt\ task's true objective for a fixed actor and critic. Then, we find the best actor and critic \wrt\ to the previously selected reward (Algorithm \ref{alg:autorl}).

For reward shaping, actor and critic network shapes are fixed sizes 
$\myvec{\theta}_{\myvec{\pi}}, \myvec{\theta_{Q}} \in I(\myvec{n}_{min}, \myvec{n}_{max})$
where $I(\myvec{a},\, \myvec{b})$ is a closed interval in n-dimensional space bounded by points $\myvec{a},\, \myvec{b} \in \R{n}.$
We run $n_{g}$ trials, at most $n_{mc}$ in parallel (Line 3, Alg. \ref{alg:hp}). 
At each trial $i$, we initialize the reward function $\myvec{\theta}_r(i)$ from $I(0,1)^n,$ based on all completed trials according to a black-box gradient-free optimization algorithm \cite{cmaes} (Line 5). The first $n_{mc}$ trials select reward weights randomly.
Next, we train asynchronous instances $RL(Actor(\myvec{\theta}_{\myvec{\pi}}), Critic(\myvec{\theta_{Q}}), R(\myvec{\theta}^i_{\myvec{r}}))$ (Line 13).
After each agent is trained, its policy is evaluated according to the true task objective Eq. \eqref{eq:obj} (Line 14).
For P2P that is Eq. \eqref{eq:p2p_obj} and for PF it is Eq. \eqref{eq:pf_obj}. 
Upon completion of all $n_{g}$ trials, the best reward (Line 17) 
\begin{equation}
\label{eq:best_reward}
\myvec{\tilde\theta}_r = \arg \max_{i}\, Obj(\myvec{\theta}_{\myvec{\pi}}, \myvec{\theta_{Q}}, \myvec{\theta}_r^i) | G)    
\end{equation}
corresponds to the trial with the highest true objective.

Next, we repeat a similar process to find the best actor and critic architecture \wrt\ to $\myvec{\tilde\theta}_r$ (Line 2, Alg. \ref{alg:autorl}). In this case, the optimization objective is to maximize the cumulative reward (Line 11, Alg. \ref{alg:hp}). This time, at each trial we train asynchronously
$RL(Actor(\myvec{\theta}^j_{\myvec{\pi}}), Critic(\myvec{\theta^j_{Q}}), R(\myvec{\tilde\theta}_{\myvec{r}}))$
and evaluate the objective \wrt Eq. \eqref{eq:reward}. 
For P2P that is Eq. \eqref{eq:p2p_reward}, and for PF it is Eq. \eqref{eq:pf_reward}.
Lastly the best actor and critic architecture
corresponds to the trial with the best return, 
\begin{equation}
\label{eq:best_hp}
\myvec{\tilde\theta}_{\myvec{\pi}},\,\myvec{\tilde\theta}_{Q} = \arg \max_j \, Obj(\myvec{\theta}_{\myvec{\pi}}^j, \myvec{\theta_{Q}}^j, \myvec{\tilde\theta}_r),    
\end{equation}
(Line 17).
The final policy trained by AutoRL is  
\begin{equation}
\label{eq:pitilde}
\myvec{\tilde\pi}(\x | W_{\myvec{\tilde\theta}_{\pi}}) = RL(Actor(\myvec{\tilde\theta}_{\myvec{\pi}}), Critic(\myvec{\tilde\theta_{Q}}), R(\myvec{\tilde\theta}_{\myvec{r}}))
\end{equation}


\begin{algorithm}[h!b]
	\caption{AutoRL} 
	\label{alg:autorl}
\begin{algorithmic}[1]
\small
\OUTPUT $\myvec{\pi}(\x | W_{\myvec{\theta}_{\myvec{\pi}}})$: Trained policy.
\STATE /* Select best reward.*/ \\
 \verb|_|$,\myvec{\tilde\theta}_{\myvec{r}}$, \verb|_|, \verb|_|$\leftarrow \text{HPSelector}(\myvec{\theta}_{\myvec{r}} = None)$ 
\STATE /* Select best NN architecture. */\\
$\myvec{\tilde\pi},$\verb|_|$,\theta^*_{\myvec{\pi}},\,\theta^*_{Q} \leftarrow \text{HPSelector}(\myvec{\tilde\theta}_{\myvec{r}})$ 
\RETURN  $\myvec{\tilde\pi}$
\end{algorithmic}
\end{algorithm}

\begin{algorithm}[h!b]
	\caption{HPSelector: Hyper-parameter selector} 
	\label{alg:hp}
\begin{algorithmic}[1]
\small
\PROPERTY $n_g$: Num. of generations (trials).
\PROPERTY $n_{mc}$: Num. of parallel trials.
\PROPERTY $\myvec{n}_{min}, \myvec{n}_{max}$: Min. and max. of neurons per layer.
\INPUT $\myvec{\theta}_{\myvec{r}}:$ Reward hyper-parameters.
\OUTPUT $\myvec{\tilde\pi}:$ Policy trained with the selected hyper-parameters.
\OUTPUT $\myvec{\tilde\theta}_{\myvec{r}}:$ Selected reward hyper-parameters.
\OUTPUT $\myvec{\tilde\theta}_{\myvec{\pi}},\,\myvec{\tilde\theta}_{Q}:$ Selected NN architecture hyper-parameters.

\STATE $T \leftarrow \emptyset$ /* Initialize the experience. */
\STATE Initialize NN architecture: $\myvec{n}_{min} \leq \myvec{\theta}_{\myvec{\pi}},\,\myvec{\theta_{Q}}
\leq\myvec{n}_{max}$ 
\FOR {$i=1,\cdots n_g$ running $n_{mc}$ trials in parallel.}
    \IF {$\myvec{\theta}_{\myvec{r}}$ is None /* Tune rewards */}
      \STATE $\myvec{\theta}_{\myvec{r}}^{i} \leftarrow$ $Select(min=0, max=1,T)$ with \cite{cmaes}
      \STATE $\myvec{\theta}^i_{\myvec{\pi}},\,\myvec{\theta^i_{Q}} \leftarrow \myvec{\theta}_{\myvec{\pi}},\,\myvec{\theta_{Q}}$ /* NN hyper-parameters are fixed.*/
      \STATE $obj_{fn} \leftarrow$ true objective (Eq. \eqref{eq:p2p_obj} or Eq. \eqref{eq:pf_obj})
    \ELSE
      \STATE $\myvec{\theta}_{\myvec{r}}^{i} \leftarrow \myvec{\theta}_{\myvec{r}}$ /* Reward hyper-parameters are fixed. */
      \STATE $\myvec{\theta}^{i}_{\myvec{\pi}},\myvec{\theta}^{i}_{Q} \leftarrow$ $Select(\myvec{n}_{min}, \myvec{n}_{max},T)$ with \cite{cmaes}
      \STATE $obj_{fn} \leftarrow$ cumulative reward.
    \ENDIF
    \STATE $ /* $ Train the agent with the selected hyper-parameters. $ */ $\\
    $\myvec{\pi}^{i} = RL(Actor(\myvec{\theta}^{i}_{\myvec{\pi}}), Critic(\myvec{\theta}^{i}_{Q}), \myvec{r}(\x, \myvec{\theta}_{\myvec{r}}^{i}))$
    \STATE $e^{i} \leftarrow$ Evaluate $\myvec{\pi}^{i}$ \wrt $obj_{fn}$ objective function.
    \STATE $T \leftarrow T \cup (e^{i}, \myvec{\pi}^{i}, \myvec{\theta}_{\myvec{r}}^{i}, \myvec{\theta}^{i}_{\myvec{\pi}}, \myvec{\theta}^{i}_{Q})$ /* Save the trial. */
  \ENDFOR
  \STATE /* Select best hyper-parameters according to Eq. \eqref{eq:best_reward} or \eqref{eq:best_hp}. */\\
  $\myvec{\tilde\pi}, \myvec{\tilde\theta}_{\myvec{r}}, \myvec{\tilde\theta}_{\myvec{\pi}},\myvec{\tilde\theta}_{Q} \leftarrow \arg max_{e \in T}\, T$
  \RETURN $\myvec{\tilde\pi}, \myvec{\tilde\theta}_{\myvec{r}}, \myvec{\tilde\theta}_{\myvec{\pi}},\myvec{\tilde\theta}_{Q}$ /* Eq. \eqref{eq:pitilde} */
\end{algorithmic}
\end{algorithm}

Algorithm \ref{alg:hp} scales linearly with the number of trials. It is important to note that the number of concurrent trials, $n_{mc},$ needs to be much smaller than the total number of trials, $n_g,$ in order for the algorithm to have enough completed experience when selecting the next set of parameters in Lines 5 and 10. If there are no completed trials, the parameters are randomly selected. Overall, the Algorithm requires $n_{mc}$ processors, and runs $\frac{n_g}{n_{mc}}$ times longer than vanilla RL. Graphical representation of the algorithm is available in Appendix A.

\section{RESULTS}
\label{sec:results}

\subsection{Setup}
\label{results:setup}

The training environment, generated from a real building floor plan, is 23m by 18m and contains no moving obstacles (Fig. \ref{fig:trainingEnv}). The simulated differential drive robot a point mass, with a 64-beam 1D lidar with 220 degrees field of view with Gaussian distributed noise $\mathcal{N}(0, \sigma_{\text{Lidar}}).$ The robot's action space is the linear and angular velocities with bounds of $[-0.2, 1.0]$ m/s and $[-1.0, 1.0]$ radian/s, respectively. 
The training and evaluation details are in Appendix B, while the details of our noise model are in the Appendix C.

Actor and critic are three-layers deep for both tasks. We choose wide and shallow feed-forward fully-connected networks, because their fast inference time makes them feasible for on-board high frequency robot control.
The critic consists of a two-layer observation embedding joined with the action network by two fully connected layers. 
We select DDPG for the learning algorithm, and Vizier \cite{vizier} with CMA-ES \cite{cmaes} for hyperparameter tuning.

We compare P2P agent with vanilla hand-tuned DDPG \cite{ddpg}, artificial potential fields (APF) \cite{khatib1986real}, Dynamic Window Approach (DWA) \cite{dwa}, and behavior cloning (BC) \cite{bc}. Behavior cloning uses the same neural network architecture as learned with AutoRL, but relies on a supervised loss instead of a reward. The PF agent is compared with guidance paths generated with PRMs \cite{kavraki-prm} with straight line local planner, and combined vanilla DDPG, APF, and DWA, to guide the robot, resulting in methods PRM-RL \cite{prm-rl}, PRM-APF \cite{chiang2015path}, and PRM-guided DWA, denoted as PRM-DWA [14*] to differentiate from DWA without a guidance path. We test the PF and P2P policies in three previously unseen large environments (Figs. \ref{fig:mtv40}, \ref{fig:sfo1ms}, \ref{fig:mtv1667}). For the P2P policy, the start and goal are randomly selected to be between 5 and 10 meters apart. For the PF, the starts and goals are randomly chosen, requiring at least 35 meters Euclidean distance between start and goal. Appendix D describes the baselines in more detail.

\subsection{Training}
\label{results:training}
We train the P2P and PF agents for 1000 trials, running 100 agents in parallel, each for 5 million training steps. Each agent takes about 12 hours to train, with the complete AutoRL run completing in several days. 
For reproducibility, Appendix D contains the parameters found by AutoRL.

\begin{figure}[tb]
\label{fig:rewards}
	\begin{center}
		\begin{tabular}{cc}
			\subfloat[\scriptsize Reward shaping generations.]{\includegraphics[width=0.22\textwidth,height=2.5cm,keepaspectratio=false]{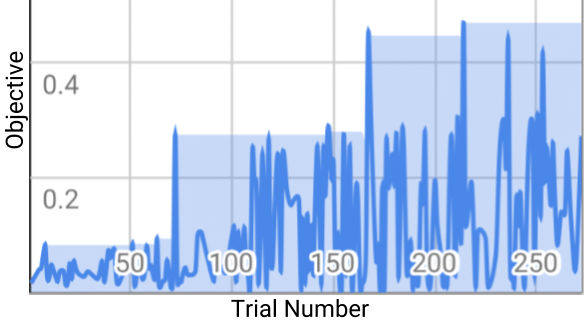}\label{fig:PFVizier}}&
			
			\subfloat[\scriptsize Reward shaping learn. curve.]{\includegraphics[width=0.22\textwidth,height=2.5cm,keepaspectratio=false]{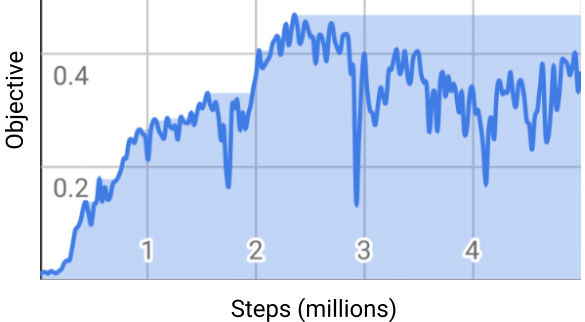}\label{fig:PFTraining}}\\
			
			\subfloat[\scriptsize Network shaping generations.]{\includegraphics[width=0.22\textwidth,height=2.5cm,keepaspectratio=false]{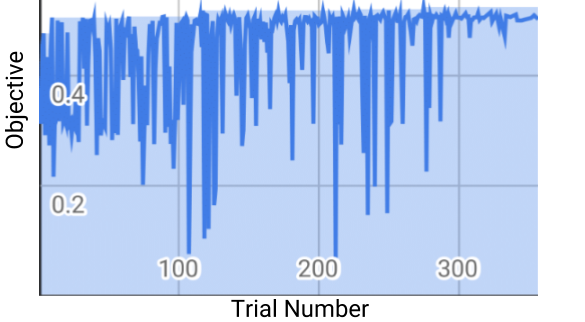}\label{fig:PFNNVizier}}&
			
			\subfloat[\scriptsize Network shaping learn. curve.]{\includegraphics[width=0.22\textwidth,height=2.5cm,keepaspectratio=false]{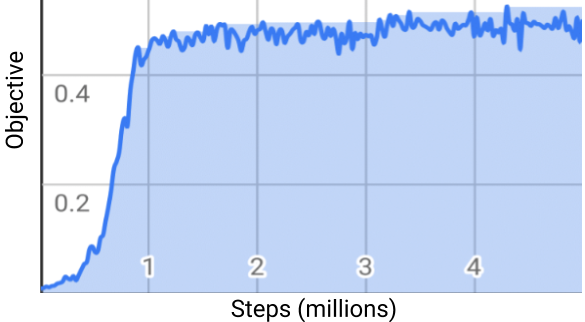}\label{fig:PFNNTraining}}
			\\
		\end{tabular}
		\caption{\footnotesize Results of AutoRL for the path following policy. Trail numbers over objective values (a) and (c). Training steps over objective (b) and (d).\label{fig:vizier}}
	\end{center}
\end{figure}

Fig. \ref{fig:vizier} shows the impact of the AutoRL on the PF task. Across both reward and network shaping there are certain parameter values where the agent does not perform well (Figs. \ref{fig:PFVizier} and \ref{fig:PFNNVizier}), although during the network shaping there are fewer bad trials. The learning curve of the non-shaped objective resembles the one depicted in Fig. \ref{fig:PFTraining}.  The best agent after reward shaping (Fig. \ref{fig:PFTraining}) shows signs on catastrophic forgetting that plagues DDPG training. Notice, that the best agent after both shaping phases (Fig. \ref{fig:PFNNTraining}) does not exhibit such forgetfulness, and the training could have been terminated sooner. 
The PF AutoRL policy achieves a true objective of 0.56, while the non-shaped policy reaches an objective of 0.26. The P2P agent trained with AutoRL has training objective of 0.90, while the hand-tuned one is 0.54. This is a promising result for the utility of AutoRL. 

\subsection{Generalization to New Environments}
\label{results:generalization}

\begin{figure}[tb]
\label{fig:generalizationPF}
\begin{tabular}{c}
		
    \subfloat[\scriptsize PF with AutoRL (Ours), RL\cite{ddpg}, PRM-APF \cite{chiang2015path}, PRM-DWA \cite{dwa}*, PRM-RL \cite{prm-rl}]{\includegraphics[trim=0mm 0mm 10mm 3mm,clip,width=0.45\textwidth,height=3.7cm,keepaspectratio=false]{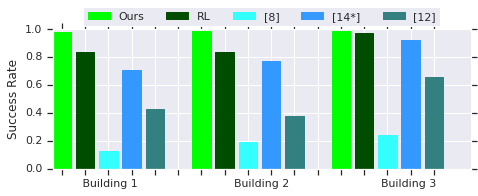}\label{fig:genSuccPF}} \\
    \subfloat[\scriptsize P2P with AutoRL (Ours), RL \cite{ddpg}, APF \cite{khatib1986real}, DWA \cite{dwa}, BC \cite{bc}]{\includegraphics[trim=0mm 0mm 10mm 3mm,clip,width=0.45\textwidth,height=3.7cm,keepaspectratio=false]{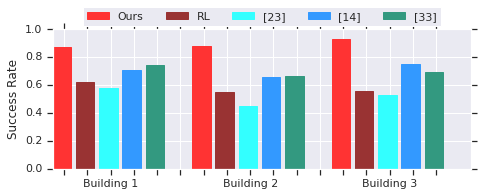}\label{fig:genSuccP2P}}
    \end{tabular}
    \caption{\footnotesize
    Success rates for PF and P2P tasks in three environments.
    }
\end{figure}

\subsubsection{PF}

AutoRL has near perfect success rate, 98.7\% on average, across all three environments (Fig. \ref{fig:genSuccPF}), and is the only method that can transfer to unseen, large, real building sized environments.
PF with AutoRL is 11\% more successful than hand tuned policy that has average success rate of 88.3\%; and 23\% more successful than best non-learned baseline (PRM-DWA with 80\% success rate). PRM-APF is consistently under performing.
The primary failure causes for both PRM-APF and PRM-DWA are wall collisions and getting stuck in a local minimum, especially when the guidance path generated by PRM is close to obstacles, which creates a local minimum \cite{chiang2015path}. PRM-RL performs worse than PRM-DWA and AutoRL, since it does not use hyperparameter tuning.

 \begin{figure}[tb]
\begin{tabular}{c}
    \subfloat[\scriptsize AutoRL]{\includegraphics[trim=5mm 95mm 5mm 35mm,clip,width=0.45\textwidth,height=2.3cm,keepaspectratio=false]{figures/trainingEnv.png}\label{fig:pfShaped}}\\
    \subfloat[\scriptsize Handed tuned DDPG]{\includegraphics[trim=5mm 95mm 5mm 39mm,clip,width=0.45\textwidth,height=2.3cm,keepaspectratio=false]{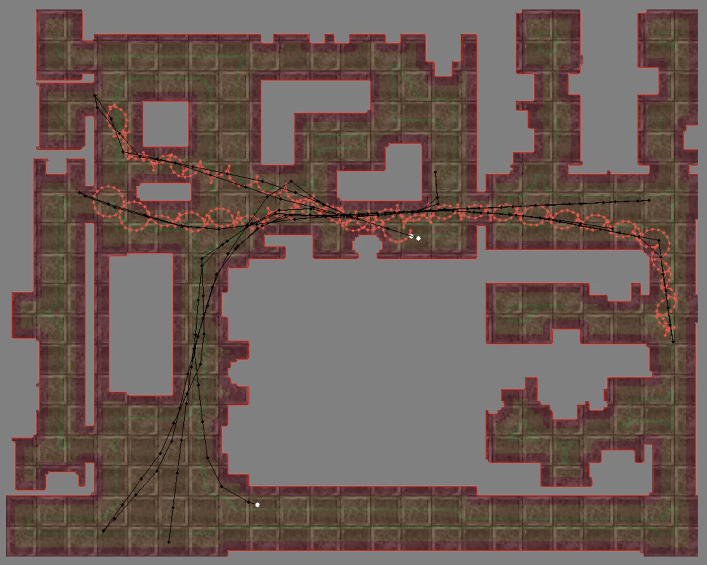}\label{fig:pfNonShaped}}
    \end{tabular}
    \caption{\footnotesize
    PF policy sample trajectories (red) and guidance paths (black) found by (a) AutoRL and (b) hand-tuned RL.\label{fig:nonShapedPF}}
\end{figure}

Curiously, the hand-tuned policy creates longer trajectories than the AutoRL one. Inspecting the trajectories (Fig. \ref{fig:nonShapedPF}), it is clear that DDPG learns completely different behaviors under the two parameterizations. 
The AutoRL policy exhibits smooth, forward-moving behavior while the hand-tuned policy alternates between forward and reverse motion, resulting in a twirling behavior with longer paths. 
Since the 220 degree field of view lidar cannot detect obstacles in the back, the twirling behavior has a lower success rate. 

\subsubsection{P2P}
Fig. \ref{fig:genSuccP2P} shows that P2P's generalization results are consistent with path following's. 
The AutoRL policy's success rate of 89\% is highest (on average 71\%, 26\%, and 27\%  higher than APF, DWA, and BC respectively) across all environments. AutoRL exhibits smooth forward motion, while the failure cases are mostly due to complex static obstacles. 
Once again, despite promising success rate of 86.5\%, the hand-tuned policy exhibits twirling, leading to high path lengths and subpar performance in noisy conditions.

\subsection{Moving Obstacle Avoidance}
\label{results:movingObstalces}

We evaluate the shaped PF and P2P policies in a large environment (Fig. \ref{fig:sfo1ms}) among up to 40 moving obstacles. The moving obstacles motion follows the social force model (SFM) \cite{helbing1995social} to mimic pedestrian motion (See Appendix C). 

\subsubsection{PF}

\begin{figure*}[]
\begin{tabular}{ccc}
    \subfloat[\scriptsize Path following]{\includegraphics[trim=0mm 0mm 0mm 3mm,clip,width=0.3\textwidth,height=3.5cm,keepaspectratio=false]{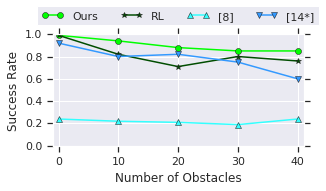}\label{fig:numSucc}} &

    \subfloat[\scriptsize 0 moving obstacles]{\includegraphics[trim=0mm 0mm 0mm 3mm,clip,width=0.3\textwidth,height=3.5cm,keepaspectratio=false]{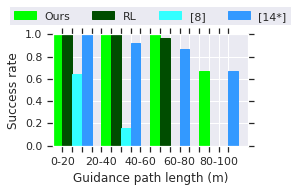}\label{fig:succVsLengthMo0}}&
    \subfloat[\scriptsize 20 moving obstacles]{\includegraphics[trim=0mm 0mm 0mm 3mm,clip,width=0.3\textwidth,height=3.5cm,keepaspectratio=false]{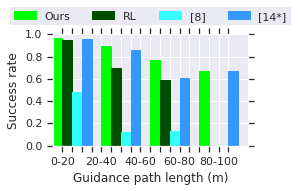}\label{fig:succVsLengthMo20}}
    \end{tabular}
    \caption{\footnotesize PF success rate over (a) number of moving obstacles, (b) guidance path length with no obstacles, and (c) guidance path length with 20 moving obstacles for AutoRL (Ours), hand tuned RL \cite{ddpg}, PRM-APF \cite{chiang2015path}, and PRM-DWA \cite{dwa}*.} \label{fig:succVsLengthPF}
\end{figure*}

AutoRL policy outperforms hand-tuned DPPG and both PRM-APF and PRM-DWA in all numbers of obstacles (Fig. \ref{fig:numSucc}). 
Note that although SFM avoids collision with the robot, simply executing the path using PRM-APF without reacting to obstacles (no repulsive potential) results in no success. This implies the robot must also partake in collision avoidance with moving obstacles. PRM-DWA can avoid moving obstacles and its performance is steady as the number of obstacles increases.

AutoRL PF performs uniformly across path guidance lengths up to 80 meters, and decreases slightly in the presence of moving obstacles, while PRM-APF degrades rapidly with guidance path length (Figs. \ref{fig:succVsLengthMo0} and \ref{fig:succVsLengthMo20}).
PRM-DWA's success is also uniform across the guidance path lengths, but is lower than AutoRL's.

\subsubsection{P2P}

\begin{figure}[]
\begin{tabular}{c}
    \subfloat[\scriptsize P2P, 0 moving obstacles]{\includegraphics[trim=0mm 0mm 0mm 3mm,clip,width=0.45\textwidth,height=3.9cm,keepaspectratio=false]{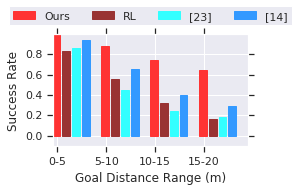}\label{fig:p2pSuccMo0}}\\
    
    \subfloat[\scriptsize P2P, 20 moving obstacles]{\includegraphics[trim=0mm 0mm 0mm 3mm,clip,width=0.45\textwidth,height=3.9cm,keepaspectratio=false]{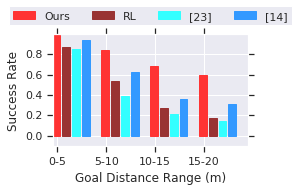}\label{fig:p2pSuccMo20}}
    \end{tabular}
    \caption{\footnotesize 
    P2P policy: AutoRL (Ours), hand tuned RL \cite{ddpg}, APF \cite{khatib1986real}, and DWA \cite{dwa} success rates (a) without and (b) with 20 moving obstacles over goal distance.\label{fig:NumObsP2P}}
\end{figure}

P2P policy's success degrades with the goal distance increase (Fig. \ref{fig:NumObsP2P}) and its success rate is not affected with the number of obstacles (Fig. \ref{fig:p2pSuccMo20}).
Hand tuned RL, APF's and DWA's performance degrade rapidly with goal distance as well, which
indicates that the main P2P failure cause, regardless of its implementation, is the inability to navigate among complex static obstacles rather than collision with moving obstacles. This is not surprising, since local planners are not designed to avoid complex static obstacles.

\subsection{Robustness to Noise}
\label{results:noise}
\begin{figure*}[tb]
\centering
    \subfloat[\scriptsize PF lidar noise]{\includegraphics[trim=0mm 0mm 0mm 3mm,clip,width=0.32\textwidth]{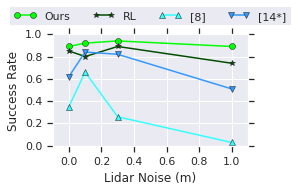}\label{fig:lidarSucc}}
    \subfloat[\scriptsize PF localization noise]{\includegraphics[trim=0mm 0mm 0mm 3mm,clip,width=0.32\textwidth]{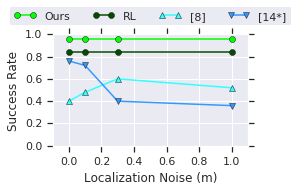}\label{fig:locSucc}}
    \subfloat[\scriptsize PF process noise]{\includegraphics[trim=0mm 0mm 0mm 3mm,clip,width=0.32\textwidth]{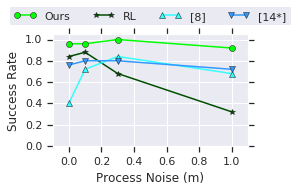}\label{fig:processSucc}}\\

    \subfloat[\scriptsize P2P lidar noise]{\includegraphics[trim=0mm 0mm 0mm 3mm,clip,width=0.32\textwidth,keepaspectratio=true]{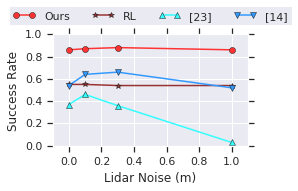}\label{fig:lidarSuccP2P}}
    \subfloat[\scriptsize P2P  localization noise]{\includegraphics[trim=0mm 0mm 0mm 3mm,clip,width=0.32\textwidth,keepaspectratio=true]{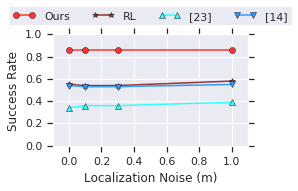}\label{fig:locSuccP2P}}
    \subfloat[\scriptsize P2P process noise]{\includegraphics[trim=0mm 0mm 0mm 3mm,clip,width=0.32\textwidth,keepaspectratio=true]{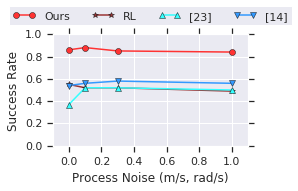}\label{fig:processSuccP2P}}
    
    \caption{\footnotesize Success rate over (left) lidar, (middle) localization noise, and (right) process noise with 20 moving obstacles. (top) Path Following compared to the hand tuned RL, PRM-APF \cite{chiang2015path} and PRM-DWA \cite{dwa}*. (bottom) P2P compared to hand tuned RL \cite{ddpg}, APF \cite{khatib1986real} and DWA \cite{dwa} navigating from random starts to goals, 5-10 m apart. \label{fig:noise}}
\end{figure*}

Fig. \ref{fig:noise} isolates one noise source at a time in order to investigate the impact on performance in an environment with 20 moving obstacles. 
Results show that PF and P2P policies are resilient to noise, even when the lidar noise is more than three times the radius of the robot ($\sigma_\text{lidar}=1$ m). 
On the other hand, lidar noise heavily influences the success of APF and DWA (for both PRM and local planner variants). This is expected since obstacle clearance is used to compute the repulsive potential for APF and compute the objective function of DWA.
In addition, the action of APF is computed by taking the gradient of the potential; such a greedy approach often guides the robot to collision or local minima.

\subsection{Physical Robot Experiments}
\label{sec:physical}

\begin{figure*}[tb]
    \centering
    \subfloat[\scriptsize Path following]{\includegraphics[trim=0mm 0mm 15mm 0mm,clip,width=0.45\textwidth,keepaspectratio=true]{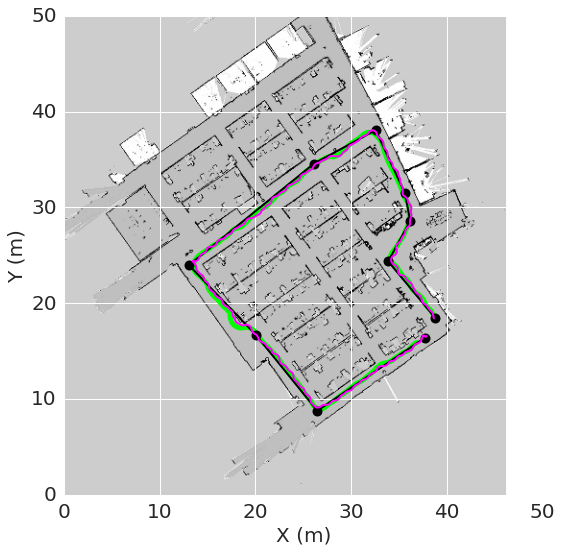}\label{fig:sim2realPF}}
    \subfloat[\scriptsize P2P]{\includegraphics[trim=0mm 0mm 15mm 0mm,clip,width=0.45\textwidth,keepaspectratio=true]{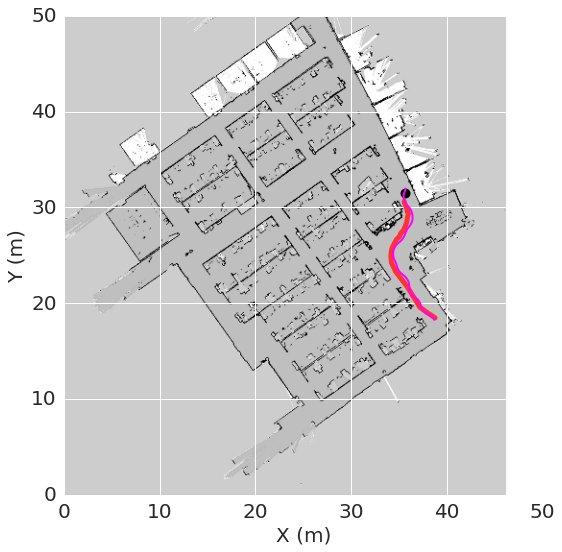}\label{fig:sim2realP2P}}
    \caption{\footnotesize (a) Path Following and (b) P2P on a real robot in an office building. The actual robot trajectories (magenta), the guidance path (black) and the trajectory in simulation (green for Path Following and red for P2P) overlaid over the 2D lidar map. \label{fig:sim2real}}
\end{figure*}

First, we investigate the difference between simulation and reality when PF and P2P policies were deployed on the Fetch robot \cite{fetch}.
For the PF task (Fig. \ref{fig:sim2realPF}), we manually specify a sequence of waypoints (black dots) as the guidance path (80.6 m in length). 
For the P2P task (Fig. \ref{fig:sim2realPF}) the start and goal are about 13.4 m apart. 
The robot reaches the goal without collision in all three runs, navigating roughly 240 m without collision.
Figs. \ref{fig:sim2realPF} and \ref{fig:sim2realP2P} (magenta) show one of robot trajectories. 
These trajectories are very close to the simulated ones (green for PF and red for P2P) but exhibit more turning than simulation, likely caused by the delay between sensing and action execution.

Next, we test the policy's performance in a very narrow corridor (only 0.3 m wider than the robot, which is 0.3 m in diameter), and with moving obstacles (Fig. \ref{fig:fetchNarrow}), qualitatively over four trials. Video at:\url{https://youtu.be/0UwkjpUEcbI} demonstrates the Fetch robot executing these policies. 
The robot reliably navigates in the corridor all four times. 
We also execute the P2P policy in a highly unstructured dynamic environment with a person playing with a dog (Fig. \ref{fig:fetchUnstructured}). The robot stops and avoids obstacles reliably. Notice in the video how the agent adapts when blocked, moves away from the goal, and around obstacles.
The only failure case we observed was caused by the obstacles undetectable by the 1D lidar such as human feet, which are below lisar's field of view.

\section{DISCUSSION}
\label{sec:dis}
AutoRL is not sample efficient: it took 12 days to train 1000 agents.
However, AutoRL learns high-quality navigation behaviors that transfer well across environments and are easy to deploy on-robot. For navigation agents, the extra training cost is justified by better quality policies. 

End-to-end learning with AutoRL effectively creates tightly-coupled perception, planning and controls.
The end-to-end POMDP setup enables robustness to noise.
The comparison with the agents on the opposite side of the spectrum, traditional and learned, highlights the benefits of end-to-end learning. The hand-tuned DDPG agent, although it produces suboptimal trajectories, is still robust to noise (\ref{fig:noise}). 
The traditional APF and DWA, not designed for tight integration between the controls and sensor, are brittle in the presence of noise.

Robustness to noise has an additional perk of overcoming guidance path imperfections. We have observed during hand-tuned training that path following is very sensitive to appropriate waypoint spacing and waypoint radius. It is only after including these parameters in reward tuning that the path-following agents learn high-quality behaviors. This is likely because the tuning finds the optimal distance from a waypoint \wrt\ to robot's noise and abilities, essentially deciding to give a credit to reaching a waypoint for what is feasible on that particular robot. 

The P2P policy found by AutoRL is more robust to local minima than APF and DWA, likely because the agent learned that following a wall usually leads to completing the goal.
As demonstrated in the video, the agent is also willing to move away from the goal to avoid local minima.
The failure point of P2P policy remains its inability to avoid large-scale local minima, such as moving from one room to another, which it was not designed to do.

\section{CONCLUSIONS}
\label{sec:conclusions}
This paper presents AutoRL, a drop-in adaptation of deep RL that automates reward and network architecture selection with large-scale hyperparameter optimization. AutoRL is used to learn two navigation building blocks, P2P and PF end-to-end behaviors. The resulting policies, although computationally expensive to train, exhibit more desirable behaviors compared to RL with hand-crafted hyperparameters and non-learned baselines.
They generalize to new environments with moving obstacles, are robust to noise and are deployed to a physical robot without tuning.
In the future, we plan to evaluate AutoRL on kinodynamic robots and high dimensional robots such as mobile manipulators.





\section*{ACKNOWLEDGMENT}
\small The authors thank J. Chase Kew, Oscar Ramirez, Lydia Tapia, Vincent Vanhoucke, and Chris Harris for helpful discussions. 

\section*{APPENDIX}
\subsection{AutoRL diagram}
\begin{figure*}[tb]
		\begin{center}
		\includegraphics[width=0.95\textwidth,keepaspectratio=true]{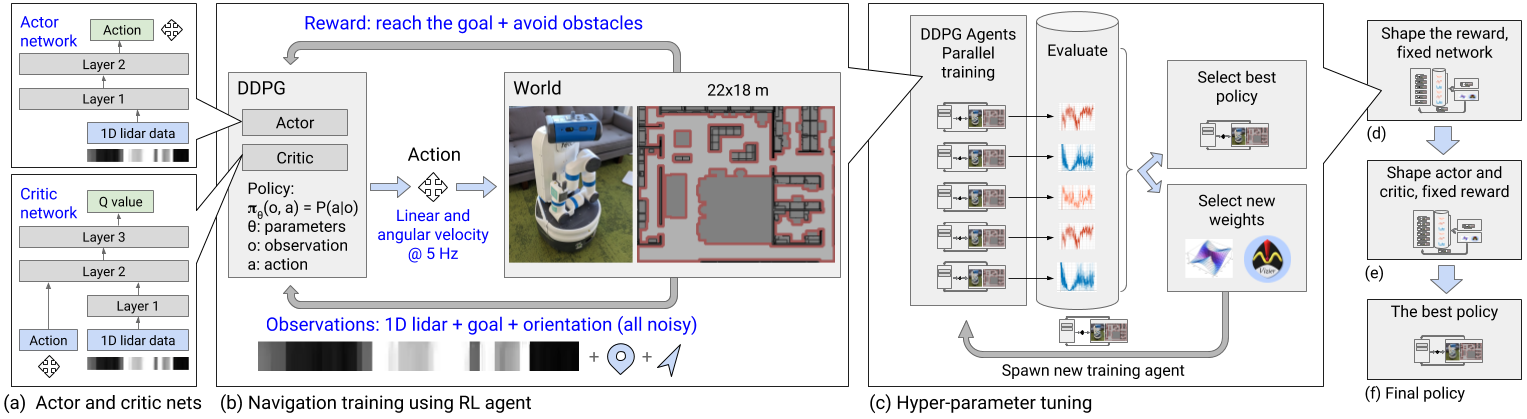}
		\caption{\small Training end-to-end policies with AutoRL. AutoRL shapes the reward function \wrt\ task's true objective with a fixed network architecture (d), then shapes the network architecture with a fixed reward function (e). AutoRL runs $n_{mc}$ parameterized vanilla DDPG agents in parallel (c). After each agent's training is complete, its performance is evaluated \wrt\ the true goal objective and stored in the database. Gradient-free optimization selects the next set of reward and network parameters, and a new agent is spawned. At the same time, the agent outputs the current best policy. Each individual agent is initialized with reward and network weights and trained with vanilla DDPG (b). The actor and critic of the vanilla DDPG are multi-layered feed-forward networks with parameterized layer widths (a). \label{fig:autorl}}
	\end{center}
\end{figure*}

Figure \ref{fig:autorl} depicts a graphical representation of the AutoRL algorithm.

\subsection{Setup}
The agent operates at $5\,$Hz. The training episodes for both P2P and PF tasks are 100 seconds long. We use TF Agents \cite{tfagents} for training. In the evaluation, we extend the episode duration to accommodate longer trajectories. The localization and orientation observations are provided by the ROS navigation stack. 

For the P2P task, the goal size is 0.25 m. 
The parameters found by AutoRL are shown in Table \ref{table:hyperparameters}.

We test the PF and P2P policies in three previously unseen large environments. All simulated evaluations are repeated 100 times. For the P2P policy the start and goal are randomly selected to be between 5 and 10 meters apart. For the PF policy, the start and goals are randomly chosen requiring at least 35 meters Euclidean distance between a start and goal.

\subsection{Robot noise modelling}
To simulate robot process noise, we added Gaussian distributed process noise $\mathcal{N}(0, \sigma_{\text{Speed}})$ and $\mathcal{N}(0, \sigma_{\text{Turning}})$ to both the linear and angular velocities. We assume the robot can localize itself but has a Gaussian distributed localization noise $\mathcal{N}(0, \sigma_{\text{Localize}})$.
Unless otherwise specified, the noise is set to $\sigma_{\text{Lidar}}=0.3$ m, $\sigma_{\text{Speed}}=0.1$ m/s, $\sigma_{\text{Turning}}=0.1$ rad/s, $\sigma_{\text{Localize}}=0.1$ m.

\begin{table}[th]
\caption{Hyperparameters Found by AutoRL}
\label{table:hyperparameters}
\begin{center}
\begin{tabular}{l|l}
\hline
P2P auto-tuned params. & Values\\
\hline
$\myvec{\theta_{r_\text{P2P}}}$ & [-0.446, 0.333, -0.120, \\
 &  0.153, 0.671, 16.081] \\
$d_{\text{P2P}}$ & 1.0 m \\
$\theta_n$ & 3\\
Actor network & [50, 20, 10] \\
Critic obs. network & [50, 20] \\
Critic joint network & [10, 10] \\
 & \\
\hline
PF auto-tuned params. & Values\\
\hline
$\myvec{\theta_{r_\text{PF}}}$ & [-0.0351, -0.90976, -34.158, \\
 &  -4.769] \\
$d_{\text{clearance}}$ & 0.5344 m\\
$d_{\text{wr}}$ & 0.371 m \\
$d_{\text{ws}}$ & 1.821 m \\
$N_\text{partial}$ & 2\\
$\theta_n$ & 4\\
Actor network & [323, 47, 560] \\
Critic obs. network & [522, 41] \\
Critic joint network & [62, 1] \\
\hline
\end{tabular}
\end{center}
\end{table}

\subsection{Baselines}
Table \ref{tab:baselines} shows the baselines used for comparison of PF anf P2P policies.

We chose APF and DWA because they are fast, well compared \cite{chiang2017safety}, and utilize only clearance information which can be obtained from 1D lidar. Many popular motion planners for dynamic environments are not suitable since they require knowledge of obstacle velocity or dynamics while we use only a 1D lidar sensor. To avoid the local minima problem that often plagues APF and DWA methods, we implemented path-guidance similar to \cite{chiang2015path}, where the attractive potential is computed along a guidance path. Guidance paths are sequences of x, y positions connecting the start to goal positions without collision. We use PRMs to generate the guidance paths in maps generated from floor plans or lidar.

The behavior cloning baseline (BC) uses the APF method for generating supervised data.
Training dataset consists of 100\,000 transitions (pairs of observations and actions) sampled from randomly generated paths of lengths 5-10 m.
Only transitions on successful paths were added to the dataset.
Test set had 10\,000 transitions.
Neural network architecture and environment settings were the same as the RL point-to-point task.
We trained for 300 epochs, each training 20 batches of size 512.
Final train accuracy on the test set was 92\%.

\begin{table*}[th]
\caption{Baselines}
\label{tab:baselines}
\begin{center}
\begin{tabular}{l|l|l|l}
\hline
Name & Citation & Baseline for & Comment \\ \hline
RL & \cite{ddpg} \cite{prm-rl} & P2P & Hand-tuned DDPG like the local planner used in \cite{prm-rl}. \\
APF & \cite{khatib1986real} & P2P & Artificial potential fields. \\
DWA & \cite{dwa} & P2P & Dynamic Window Approach (DWA) . \\
BC & \cite{bc} & P2P & Behavior cloning. Same neural network architecture and environment. \\
RL & \cite{ddpg} & PF & Same as the AutoRL PF, buth with the hand-selected parameters. \\
PRM-APF & \cite{chiang2015path} & PF & Guidance path provided by PRMs, and APF follows the path. \\
PRM-DWA & \cite{dwa}* & PF & Guidance path provided by PRMs, and DWA follows the path. \\
PRM-RL & \cite{prm-rl} & PF & Guidance path provided by PRMs, and RL follows the path between the waypoints. \\
\hline

\hline
\end{tabular}
\end{center}
\end{table*}

\subsection{Moving obstacles setup}
We use a path guidance similar to \cite{chiang2015path} to create a guidance path for each moving obstacle and compute the desired velocity component in Social Force Model (SFM), alleviating the SFM limitation of moving obstacles getting trapped in local minima \cite{reif1999social}.

\bibliographystyle{abbrv}

\bibliography{literature}

\end{document}